# Convex Subspace Clustering by Adaptive Block Diagonal Representation

Yunxia Lin and Songcan Chen, *Member, IEEE*

*Abstract*—Subspace clustering is a class of extensively studied clustering methods where the spectral-type approaches are its important subclass. Its key first step is to desire learning a representation coefficient matrix with block diagonal structure. To realize this step, many methods were successively proposed by imposing different structure priors on the coefficient matrix. These impositions can be roughly divided into two categories, i.e., indirect and direct. The former introduces the priors such as sparsity and low rankness to indirectly or implicitly learn the block diagonal structure. However, the desired block diagonalty cannot necessarily be guaranteed for noisy data. While the latter directly or explicitly imposes the block diagonal structure prior such as block diagonal representation (BDR) to ensure so-desired block diagonalty even if the data is noisy but at the expense of losing the convexity that the former's objective possesses. For compensating their respective shortcomings, in this article, we follow the direct line to propose adaptive BDR (ABDR) which explicitly pursues block diagonalty without sacrificing the convexity of the indirect one. Specifically, inspired by Convex BiClustering, ABDR coercively fuses both columns and rows of the coefficient matrix via a specially designed convex regularizer, thus naturally enjoying their merits and adaptively obtaining the number of blocks. Finally, experimental results on synthetic and real benchmarks demonstrate the superiority of ABDR to the state-of-the-arts (SOTAs).

## I. INTRODUCTION

SUBSPACE clustering is a class of classic clustering methods to deal with high-dimensional data such as image, video, text, and has been extensively studied so far [1]–[9]. It assumes that given high-dimensional data points are approximately drawn from a union of some low-dimensional subspaces and aims to segment them into corresponding subspaces as faithfully as possible [4], [10]. The research on this topic has fostered various applications in, for example, machine learning [11], computer vision [12], image processing [13], [14],
and system identification [15]. Alternatively, various subspace clustering methods are proposed and mainly cover four types: iterative method [16], [17], algebraic method [18]–[20], statistical method [21]–[23], and spectral-type method [24]–[26]. Among them, the spectral-type method has almost become the most attractive and popular one in recent years due to its simplicity and excellent performance [27], [28]. The approaches of this type usually perform the following two steps: the first step constructs or learns the representation coefficients from the collected data for constructing corresponding Laplacian or affinity matrix and the second step performs spectral clustering based on the Laplacian matrix to determine the final partition/segmentation. Here, learning the representation coefficients in the first step plays a key role for the clustering effectiveness. However, owing to the complexity and diversity of the inherently unknown structure of the real data, we have to introduce some assumptions on data distributions, such as manifold or low rank assumption in the representation learning, thus leading to different ways to construct or learn representation coefficients. These ways can further be subdivided into two categories: locality-inducing method and globality-inducing method.

The locality-inducing method directly defines representation coefficient for each data point by using multiple samples in its neighborhood according to certain distance metric or proximity. For example, $\varepsilon$-ball neighborhood, $k$-nearest neighbor (KNN) [24], local subspace affinity (LSA) [29]. Actually, the data points drawn from the union of multiple low-dimensional subspaces may be distributed arbitrarily rather than locally. As a consequence, this locality-inducing manner is not enough to reflect the whole subspace structure of data.

The globality-inducing method exploits the global subspace structure assumption to learn the representation coefficients, in which the most typical and commonly adopted one is self-expression based, i.e., each data in the union of some low-dimensional subspaces can be represented as a linear combination of other data points [30]. That is $\mathbf{X} = \mathbf{XZ}$, where $\mathbf{X} = [x_1, x_2, \ldots, x_n] \in \mathcal{R}^{d \times n}$ is the given data matrix and $\mathbf{Z} = [z_1, z_2, \ldots, z_n] \in \mathcal{R}^{n \times n}$ is the self-expression coefficient matrix to be learned, in which $d$ and $n$ are the number of dimension and the number of data points, respectively. Essentially, the approaches of this category aim to automatically represent a data point as a linear combination of the data points from the target subspace while make the representation coefficients corresponding to the data points from the nontarget subspaces almost zero. Note that in this

Manuscript received March 18, 2021; revised December 10, 2021; accepted March 24, 2022. This work was supported by the Key Program of National Natural and Science Foundation of China (NSFC) under Grant 61732006. *(Corresponding author: Songcan Chen.)*

The authors are with the College of Computer Science and Technology/College of Artificial Intelligence, Nanjing University of Aeronautics and Astronautics, Nanjing 211106, China, and also with the MIIT Key Laboratory of Pattern Analysis and Machine Intelligence, Nanjing 211106, China (e-mail: linyx@nuaa.edu.cn; s.chen@nuaa.edu.cn).

This article has supplementary material provided by the authors and color versions of one or more figures available at https://doi.org/10.1109/TNNLS.2022.3164540.

Digital Object Identifier 10.1109/TNNLS.2022.3164540

case, $\mathbf{Z}$ forms the desirable block diagonal structure, thus revealing the true membership of the data matrix $\mathbf{X}$. If we next apply spectral clustering on the affinity matrix constructed by such $\mathbf{Z}$, we are very likely to acquire desired clustering. Therefore, the block diagonal matrix plays a vital role in spectral-type subspace clustering. To obtain such near-true block diagonal representation (BDR) matrix, various structure priors are introduced by imposing corresponding regularizers on $\mathbf{Z}$, fostering respective block diagonalty-oriented formulations. These formulations roughly cover two learning categories, i.e., indirect and direct. Among them, the indirect category introduces some structure priors induced by different regularizations on $\mathbf{Z}$ to indirectly learn the block diagonal structure. For example, sparse subspace clustering (SSC) [30], [31] introduces sparsity on $\mathbf{Z}$ induced by $\ell_1$ norm. Low rank representation (LRR) [32], [33] imposes low rankness on $\mathbf{Z}$ induced by nuclear norm. Multisubspace representation (MSR) [34] imposes both sparsity and low rankness on $\mathbf{Z}$ by $\ell_1$ norm and nuclear norm. Least squares regression (LSR) [35] enforces high correlation on $\mathbf{Z}$ formulated by $F$ norm. Subspace segmentation via quadratic programming (SSQP) [36] minimizes $||\mathbf{Z}^T \mathbf{Z}||_1$ with nonnegative $\mathbf{Z}$. However, these methods can obtain the desired block diagonality of $\mathbf{Z}$ only when the data points are clean, implying that such block diagonalty cannot necessarily be guaranteed in real applications due to the noise or corruption in data. While the direct category directly introduces the block diagonal structure prior on $\mathbf{Z}$ to ensure the so-desired block diagonalty even if the data is noisy. For example, the approach [37] and BDR [27] directly enforce the block diagonal structure of $\mathbf{Z}$ imposed by a hard graph Laplacian constraint and a soft self-defined block diagonal matrix induced regularizer, respectively. But it is worthy to note that the superior performance of such category is obtained at the expense of sacrificing the convexity of the indirect one, while the nonconvexity also easily gets stuck into local minima. Additionally, different from the indirect methods, the number of subspaces in such categorical approach needs to be prefixed.

In this work, to compensate the respective shortcomings of the above two categories, we follow the direct way to propose adaptive BDR (ABDR) which explicitly pursues the block diagonalty of the coefficient matrix without losing the convexity of indirect way. Specifically, inspired by Convex BiClustering [38], we enforce the columns and the rows of the coefficient matrix to be simultaneously shrunk by adding corresponding convex auto-fused terms into the squared data-fidelity term, thus adaptively identifying the groups of rows and the groups of columns in the coefficient matrix that are associated with each other. That is, the block diagonal pattern automatically emerges when the rows and columns are fused to certain extent without imposing any structure prior and prefixing the number of blocks. In summary, the contributions of ABDR can be summarized as follows.

1) ABDR can adaptively form the desirable block structure of the coefficient matrix without imposing any structure prior on it.
2) ABDR can be established from a well-defined convex objective function where generalized alternating direction method of multiplier (GADMM) can be used to achieve efficient optimization.
3) ABDR only involves one hyperparameter, making its adjustment relatively easier.
4) ABDR can straightforwardly be extended to solve various problems in subspace clustering, such as deep ABDR and kernel ABDR for nonlinear data.
5) Our experimental results validate the effectiveness and efficiency of our proposed ABDR on three benchmark subspace clustering tasks compared with several state-of-the-art (SOTA) spectral-type subspace clustering methods.

Table I compares existing spectral-type subspace clustering methods with our proposed method ABDR in main aspects.

In the rest of this article, Section II briefly overviews the related works including SSC, LRR and BDR. Section III details our algorithm (ABDR). Section IV reports extensive experimental results and analysis. Finally, Section V concludes this article with future research directions.

## II. RELATED WORK

In this section, we review three SOTAs of self-expression based spectral-type subspace clustering methods consisting of two indirect ones, i.e., SSC, LRR [39], [40], and one direct method, i.e., BDR [27].

Before detailing the related work, it is necessary to formulate the self-expression property and block diagonalty. Assume that we have a data matrix $\mathbf{X} \in \mathcal{R}^{d \times n}$ sampled from a union of $k$ subspaces $\{\mathbf{S}_i\}_{i=1}^k$ at hand. According to the subspace structure, the sampled data points follow the *self-expression property* that each data point in the union of subspaces can be represented by a linear combination of other data points, that is $x_i = \mathbf{X} z_i$. This can be formulated as a matrix form $\mathbf{X} = \mathbf{X}\mathbf{Z}$, where $\mathbf{Z} \in \mathcal{R}^{n \times n}$ is the representation coefficient matrix. Assume each subspace $\mathbf{S}_i$ has $n_i$ data points with $\sum_{i=1}^k n_i = n$, let $\mathbf{X}_i$ is a subdataset constructed by these data points. We can rewrite $\mathbf{X} = \{\mathbf{X}_1, \mathbf{X}_2, \ldots, \mathbf{X}_k\}$. Essentially, we aim to obtain a representation coefficient matrix $\mathbf{Z}$ such that each data point is represented as the linear combination of data points only from the target subspace, i.e., $\mathbf{X}_i = \mathbf{X}_i \mathbf{Z}_i$. In this case, $\mathbf{Z}$ has the *k-block diagonal structure*, that is

$$\mathbf{Z} = \begin{pmatrix} \mathbf{Z}_1 & 0 & \cdots & 0 \\ 0 & \mathbf{Z}_2 & \cdots & 0 \\ \cdots & \cdots & \ddots & \cdots \\ 0 & 0 & 0 & \mathbf{Z}_k \end{pmatrix}, \quad \mathbf{Z}_i \in \mathcal{R}^{n_i \times n_i}. \qquad (1)$$

If $\mathbf{Z}$ has $k$-block diagonal structure, we say that $\mathbf{Z}$ obeys *block diagonalty*.

### A. Sparse Subspace Clustering

To obtain the desirable block diagonal structure of the coefficient matrix, SSC [30], [31] introduces sparsity on $\mathbf{Z}$ by regularizing $\mathbf{Z}$ with $\ell_1$ norm. While considering $\mathbf{X} = \mathbf{X}\mathbf{Z}$ may not hold in real-world noisy data, SSC finally proposes to optimize the following objective:

$$\min_{\mathbf{Z}} ||\mathbf{Z}||_1 + \lambda ||\mathbf{X} - \mathbf{X}\mathbf{Z}||_F^2, \quad \text{s.t. } \text{diag}(\mathbf{Z}) = 0. \qquad (2)$$

TABLE I
COMPARISON BETWEEN MULTIPLE SUBSPACE CLUSTERING METHODS

|  | Method | Regularization | Constraint | # of Parameters | Assumption | Convexity | Optimization |
|---|---|---|---|---|---|---|---|
| Locality | $\epsilon$-neighborhood [24] | - | - | - | - | - | - |
|  | KNN [24] | - | - | - | - | - | - |
|  | LSA [29] | - | - | - | - | - | - |
| Globality | SSC [30][31] | $\|\|\mathbf{Z}\|\|_1$ | ✓ | 1 | ✓ | ✓ | ADMM |
|  | LRR [32][33] | $\|\|\mathbf{Z}\|\|_*$ | - | 1 | ✓ | ✓ | ADMM |
|  | MSR [34] | $\|\|\mathbf{Z}\|\|_1 + \lambda\|\|\mathbf{Z}\|\|_*$ | ✓ | 1 | ✓ | ✓ | ALM |
|  | LSR [35] | $\|\|\mathbf{Z}\|\|^2$ | - | 1 | ✓ | ✓ | closed solution |
|  | SSQP [36] | $\|\|\mathbf{Z}^T\mathbf{Z}\|\|_1$ | ✓ | 1 | ✓ | ✓ | SPG |
|  | HBDR [37] | $\|\|\mathbf{Z}\|\|_1$ OR $\|\|\mathbf{Z}\|\|_*$ | ✓ | 1 | - | × | ALM |
|  | BDR [27] | $\|\|\mathbf{B}\|\|_{\overline{k}} = \sum_{i=N-k+1}^N \lambda_i(L_\mathbf{B})$ | ✓ | 2 | - | × | ALM |
| Mixture | ABDR | $\Omega(\mathbf{Z})$ | - | 1 | - | ✓ | GADMM |

- **Constraint:** "✓" imposes restriction on **Z**; "-" indicates no restriction on **Z**.
- **Assumption:** "✓" indicates certain subspace assumption required; "-" indicates no subspace assumption required.
- **Convexity:** "✓" indicates the objective function is convex; "×" indicates the objective function is non-convex.
- **Optimization:** "-" indicates no need to optimize; "ADMM" represents **A**lternating **D**irection **M**ethod of **M**ultipliers; "ALM" represents **Al**ternating **M**inimization; "SPG" represents **S**pectral **P**rojected **G**radient method; "GADMM" represents **G**eneralized **A**lternating **D**irection **M**ethod of **M**ultipliers.

It has been proven that SSC can acquire the block diagonal coefficient matrix when the multiple low-dimensional subspaces are mutually independent [30], [31]. However, the data points may not strictly lie on the independent subspaces in real applications, thus the block diagonal structure of **Z** obtained by SSC is hard to be guaranteed practically.

### B. Low Rank Representation

To determine the block diagonal coefficient matrix, LRR [32], [33] introduces low rankness on **Z** by regularizing **Z** with nuclear norm. Considering $\mathbf{X} = \mathbf{XZ}$ may not strictly hold for real-world noisy data, LRR instead optimizes the following objective:

$$\min_\mathbf{Z} ||\mathbf{Z}||_* + \lambda ||\mathbf{X} - \mathbf{XZ}||_F^2. \quad (3)$$

Liu et al. [32], [33] prove that the **Z** obtained by LRR can be block diagonal only over ideal independent subspaces. However, for real application settings, the block diagonality of **Z** obtained by LRR is not largely ensured.

### C. Block Diagonal Representation

To obtain the desirable block diagonal coefficient matrix even for noisy data, BDR [27] directly enforces the block diagonal structure of the coefficient matrix **Z** by imposing a specially designed soft regularizer on **Z**, and then defines and optimizes the following objective:

$$\min_{\mathbf{B},\mathbf{Z}} \frac{1}{2}||\mathbf{X} - \mathbf{XZ}||^2 + \frac{\lambda}{2}||\mathbf{Z} - \mathbf{B}||^2 + \gamma ||\mathbf{B}||_{\overline{k}}$$
$$\text{s.t. } \text{diag}(\mathbf{B}) = 0, \quad \mathbf{B} \geq 0, \quad \mathbf{B} = \mathbf{B}^T \quad (4)$$

where $||\mathbf{B}||_{\overline{k}} = \sum_{i=n-k+1}^n \lambda_i(\mathbf{L_B})$ and $\mathbf{L_B}$ is the Laplacian matrix of **B**. Although BDR can obtain relatively more desirable block diagonal coefficient matrix for noisy data, it not only sacrifices the convexity of SSC and LRR, but also introduces an additional hyperparameter in its model compared with SSC, LRR, and ours, making its determination more complicated. Moreover, different from SSC, LRR, and ours, the number of subspaces in BDR must be given in advance.

## III. CONVEX SUBSPACE CLUSTERING BY ABDR

In this section, we first detail Convex Subspace Clustering by ABDR and then provide its problem solution including optimization strategy, complexity analysis, and convergence analysis. Finally, we describe the subspace clustering algorithm.

### A. Model Formulation

Different from the traditional clustering which only focuses on clustering the samples (columns), the biclustering puts the columns and the rows on equal footing and simultaneously groups samples (columns) and features (rows) in a data matrix [41] to uncover structures in both the column and row variables. Among many biclustering methods, the Convex BiClustering [38], a newborn and charming biclustering method with the goal to identify the groups of columns and groups of rows that are associated with each other, outperforms as the first convex attempt. To realize this goal, it simultaneously fuses the columns and the rows in data matrix via a self-designed convex regularizer and thus drives the desired pattern automatically emerge. Inspired by such method, we incorporate the corresponding auto-fused terms into the data fitting term to simultaneously shrink the columns and rows of the representation matrix and formulate the problem (5) of Convex Subspace Clustering by ABDR

$$\min_\mathbf{Z} \frac{1}{2}||\mathbf{X} - \mathbf{XZ}||_F^2 + \gamma \, \Omega(\mathbf{Z}) \quad (5)$$

where $\Omega(\mathbf{Z}) = \sum_{(i,j)\in E} w_{ij}||\mathbf{Z}_{\cdot i} - \mathbf{Z}_{\cdot j}||_2 + \sum_{(i,j)\in E} w_{ij}||\mathbf{Z}_{i\cdot} - \mathbf{Z}_{j\cdot}||_2$ and is convex in **Z**, $E$ is the edge set formed by the point-pairs of data and $\mathbf{Z}_{\cdot i}$ ($\mathbf{Z}_{i\cdot}$) indicates the $i$th column (row) of the coefficient matrix **Z** and $w_{ij}$ is a nonnegative weight indicating the similarity between two data points $\mathbf{X}_{\cdot i}$ and $\mathbf{X}_{\cdot j}$. Because both each column and its corresponding row in **Z** are the new representations of original sample, so the column fused term and the row fused term should share the same weight $w_{ij}$. Similar with [48], we use the sparse Gaussian kernel weights in (5), i.e., $w_{ij} = \iota_{(i,j)}^k \exp(-\phi ||\mathbf{X}_{\cdot i} - \mathbf{X}_{\cdot j}||_2^2)$. Here, $\iota_{(i,j)}^k$ is an indicator

function whose value is 1 if $\mathbf{X}_{.j}$ is among the KNNs of $\mathbf{X}_{.i}$ or 0 otherwise. $\exp(-\phi\|\mathbf{X}_{.i} - \mathbf{X}_{.j}\|_2^2)$ encourages the columns and the rows corresponding to the more similar samples to fuse. The nonnegative constant $\phi$ controls the rate of pressure on the fusion. The value $\phi = 0$ corresponds to uniform weights. Note that the objective function (5) of ABDR is convex in $\mathbf{Z}$, thus ensuring its global minimizer. Here, it is necessary to emphasize a major difference between ABDR and the convex biclustering [38]. Different from convex biclustering which aims to automatically group or partition given dataset in the sample and feature ways in the original space, what ABDR does is to adaptively group columns and rows of new representations in the new representation space.

When the parameter $\gamma = 0$, $\mathbf{Z}$ is an identity matrix where no row and column is associated. As $\gamma$ increases, the groups of rows and the groups of columns that are associated are simultaneously shrunk toward each other. That is ABDR identifies the groups of columns and groups of rows of $Z$ that correspond to the same subspace while automatically makes the groups of columns and groups of rows of $Z$ corresponding to different subspaces approximate to zero. When $\gamma$ is large to some extent, the desired block diagonal pattern automatically emerges in the coefficient matrix without imposing any structure prior and prefixing the number of clusters/subspaces.

**Theorem 1** further verifies the block diagonalty of $\mathbf{Z}$ obtained by ABDR. The detailed proof of **Theorem 1** can be found in the supplementary material.

*Theorem 1:* Assume a set of data points drawn from $k$ independent subspaces $\{S_i\}_{i=1}^k$. Let $\mathbf{X}_i \in \mathcal{R}^{n_i \times n_i}$ define the subset of data points sampled from $S_i$ with rank of $d_i$ and $\sum_{i=1}^k n_i = n$. Let $\mathbf{X} = [\mathbf{X}_1, \mathbf{X}_2, \ldots, \mathbf{X}_k] \in \Delta$, where $\Delta$ is a set of matrices with nonzero columns. Let $\mathbf{Z}^*$ is any optimal solution of (7), then $\mathbf{Z}^*$ satisfies the block diagonal property, that is

$$\mathbf{Z}^* = \begin{pmatrix} \mathbf{Z}_1 & 0 & \cdots & 0 \\ 0 & \mathbf{Z}_2 & \cdots & 0 \\ \cdots & \cdots & \ddots & \cdots \\ 0 & 0 & 0 & \mathbf{Z}_k \end{pmatrix} \quad (6)$$

with each $\mathbf{Z}_i \in \mathcal{R}^{n_i \times n_i}$ corresponding to $\mathbf{X}_i$

$$\min_{\mathbf{Z}} \Omega(\mathbf{Z})$$
$$\text{s.t.} \quad \mathbf{X} = \mathbf{XZ} \quad (7)$$

where $\Omega(\mathbf{Z}) = \sum_{(i,j) \in E} \mathbf{w}_{ij} \|\mathbf{Z}_{.i} - \mathbf{Z}_{.j}\|_2 + \sum_{(i,j) \in E} \mathbf{w}_{ij} \|\mathbf{Z}_{i.} - \mathbf{Z}_{j.}\|_2$.

### B. Problem Solution

We adopt the GADMMs [42], [43] to optimize our proposed ABDR. Before detailing the optimization procedure, we first give some preliminaries and notations.

*1) Preliminaries and Notations:* Following [47], for a given undirected graph $G = (V, E)$ where $V$ is the set of $n$ vertices and $E$ is the set of edges, according to the enumeration order of the index pairs in $E$, which is denoted by $l(i, j)$ for the point-pair $(i, j)$, we give the node-arc incidence matrices $\mathcal{J}$, $\widetilde{\mathcal{J}}$ as follows:

$$\mathcal{J}_k^{l(i,j)} = \begin{cases} 1, & \text{if } k = i \\ 0, & \text{otherwise}; \end{cases} \quad \widetilde{\mathcal{J}}_k^{l(i,j)} = \begin{cases} 1, & \text{if } k = j \\ 0, & \text{otherwise} \end{cases}$$

where $\mathcal{J}_k^{l(i,j)}$, $\widetilde{\mathcal{J}}_k^{l(i,j)}$ are the $k$th entry of the $l(i, j)$th column of $\mathcal{J} \in \mathcal{R}^{n \times |E|}$ and the $k$th entry of the $l(i, j)$th column of $\widetilde{\mathcal{J}} \in \mathcal{R}^{n \times |E|}$, respectively, where $|E|$ is the number of edges in $G$. For the given coefficient matrix $\mathbf{Z} \in \mathcal{R}^{n \times n}$ and the graph $G$, let $\mathcal{Q} = \mathcal{J} - \widetilde{\mathcal{J}}$, we can write column difference matrix of coefficient matrix as $\mathcal{B}_{\text{col}}(\mathbf{Z}) = \mathbf{Z}\mathcal{Q}$ and row difference matrix of coefficient matrix $\mathbf{Z}$ as $\mathcal{B}_{\text{row}}(\mathbf{Z}) = \mathcal{Q}^T \mathbf{Z}$, respectively, where $\mathcal{B}_{\text{col}} : \mathcal{R}^{n \times n} \to \mathcal{R}^{n \times |E|}$ and $\mathcal{B}_{\text{row}} : \mathcal{R}^{n \times n} \to \mathcal{R}^{|E| \times n}$ are two linear operators. Therefore, the formulation (5) can be rewritten as

$$\min_{\mathbf{Z}} \frac{1}{2}\|\mathbf{X} - \mathbf{XZ}\|_F^2 + \gamma \left([\mathcal{B}_{\text{col}}(\mathbf{Z})]_{\text{col},2} \mathbf{w} + \mathbf{w}^T [\mathcal{B}_{\text{row}}(\mathbf{Z})]_{\text{row},2}\right) \quad (8)$$

where $\mathbf{w}$ is an edge weight column vector, among which the order of point-pair $(i, j)$ is the same as that in $E$. Besides, for a matrix $\mathbf{M} \in \mathcal{R}^{p \times q}$, we define $[\mathbf{M}]_{\text{col},2} = [\|\mathbf{M}_{.1}\|_2, \ldots, \|\mathbf{M}_{.q}\|_2]$ and $[\mathbf{M}]_{\text{row},2} = [\|\mathbf{M}_{1.}\|_2, \ldots, \|\mathbf{M}_{p.}\|_2]^T$.

*2) Optimization Algorithm:* We first convert the formulation (8) to the following equivalent problem (9) by introducing two auxiliary variables $\mathbf{V}_1$ and $\mathbf{V}_2$:

$$\min_{\mathbf{Z}} \frac{1}{2}\|\mathbf{X} - \mathbf{XZ}\|_F^2 + \gamma \left([\mathbf{V}_1]_{\text{col},2} \mathbf{w} + \mathbf{w}^T [\mathbf{V}_2]_{\text{row},2}\right)$$
$$\text{s.t.} \quad \mathcal{B}_{\text{col}}(\mathbf{Z}) = \mathbf{V}_1$$
$$\mathcal{B}_{\text{row}}(\mathbf{Z}) = \mathbf{V}_2 \quad (9)$$

which can be solved by solving the following augmented Lagrange multipliers (ALMs) problem:

$$\min_{\mathbf{Z},\mathbf{V}_1,\mathbf{V}_2,\Lambda,\Psi} \frac{1}{2}\|\mathbf{X} - \mathbf{XZ}\|_F^2 + \gamma \left([\mathbf{V}_1]_{\text{col},2} \mathbf{w} + \mathbf{w}^T [\mathbf{V}_2]_{\text{row},2}\right)$$
$$+ \text{tr}\left(\Lambda^T (\mathcal{B}_{\text{col}}(\mathbf{Z}) - \mathbf{V}_1)\right)$$
$$+ \text{tr}\left(\Psi^T (\mathcal{B}_{\text{row}}(\mathbf{Z}) - \mathbf{V}_2)\right)$$
$$+ \frac{\mu_1}{2}\|\mathcal{B}_{\text{col}}(\mathbf{Z}) - \mathbf{V}_1\|_F^2 + \frac{\mu_2}{2}\|\mathcal{B}_{\text{row}}(\mathbf{Z}) - \mathbf{V}_2\|_F^2 \quad (10)$$

where both $\Lambda \in \mathcal{R}^{n \times |E|}$ and $\Psi \in \mathcal{R}^{|E| \times n}$ are Lagrange multipliers and $\mu_1, \mu_2 > 0$ are two penalty parameters. Here, we adopt the GADMMs [42], [43] to solve (10), which is far more efficient for large problems [43].

*Step 1 (Fix Others, Update $\mathbf{Z}$):* To update $\mathbf{Z}$, we need to optimize the following function:

$$f(\mathbf{Z}) = \frac{1}{2}\|\mathbf{X} - \mathbf{XZ}\|_F^2 + \text{tr}(\Lambda^T (\mathcal{B}_{\text{col}}(\mathbf{Z}) - \mathbf{V}_1))$$
$$+ \text{tr}(\Psi^T (\mathcal{B}_{\text{row}}(\mathbf{Z}) - \mathbf{V}_2)) + \frac{\mu_1}{2}\|\mathcal{B}_{\text{col}}(\mathbf{Z}) - \mathbf{V}_1\|_F^2$$
$$+ \frac{\mu_2}{2}\|\mathcal{B}_{\text{row}}(\mathbf{Z}) - \mathbf{V}_2\|_F^2 \quad (11)$$

let $\widetilde{\mathbf{V}}_1 = \mathbf{V}_1 + (1/\mu_1)\Lambda$, $\widetilde{\mathbf{V}}_2 = \mathbf{V}_2 + (1/\mu_2)\Psi$, (11) can be reformulated as

$$f(\mathbf{Z}) = \frac{1}{2}||\mathbf{X} - \mathbf{XZ}||_F^2 + \frac{\mu_1}{2}||\mathcal{B}_{\text{col}}(\mathbf{Z}) - \widetilde{\mathbf{V}}_1||_F^2 \\ + \frac{\mu_2}{2}||\mathcal{B}_{\text{row}}(\mathbf{Z}) - \widetilde{\mathbf{V}}_2||_F^2. \quad (12)$$

For highly efficient $\mathbf{Z}$-update, following [43], we augment $\mathbf{Z}$-update with the quadratic operator $\Xi(\mathbf{Z}) = (1/2)(\alpha||\mathbf{Z}||_F^2 - \mu_1||\mathcal{B}_{\text{col}}(\mathbf{Z})||_F^2 - \mu_2||\mathcal{B}_{\text{row}}(\mathbf{Z})||_F^2)$. That is

$$\mathbf{Z} = \arg\min f(\mathbf{Z}) + \Xi(\mathbf{Z} - \mathbf{Z}^k) \quad (13)$$

where $\mathbf{Z}^k$ is $\mathbf{Z}$ obtained from the previous iteration. Note that $\alpha$ must be chosen to guarantee that $\Xi(\mathbf{Z})$ is positive definite. In fact, [43] has given the smallest valid value of $\alpha$ where $\alpha > 2\rho(\sigma(\mathcal{Q}))$. Here, $\sigma(\mathcal{Q})$ is the largest singular value of matrix $\mathcal{Q}$.

Let the gradient of $f(\mathbf{Z}) + \Xi(\mathbf{Z} - \mathbf{Z}^k)$ with respect to $\mathbf{Z}$ be zero, we can easily obtain the $\mathbf{Z}$-update in each iteration is

$$\mathbf{Z} = (\mathbf{X}^T\mathbf{X} + \alpha\mathbf{I})^{-1}(\alpha\mathbf{Z}^k + \mathbf{X}^T\mathbf{X} + \mu_1\mathbf{A} + \mu_2\mathbf{B}) \quad (14)$$

where $\mathbf{I}$ is an identity matrix, $\mathbf{A} = (\mathbf{V}_1 + (1/\mu_1)\Lambda - \mathbf{Z}^k\mathcal{Q})\mathcal{Q}^T$, $\mathbf{B} = \mathcal{Q}(\mathbf{V}_2 + (1/\mu_2)\Psi - \mathbf{Z}^k\mathcal{Q}^T)$.

*Step 2 (Fix Others, Update $\mathbf{V}_1$):* To update $\mathbf{V}_1$, we need to optimize the following function:

$$f(\mathbf{V_1}) = \gamma[\mathbf{V}_1]_{\text{col},2}\mathbf{w} + \text{tr}(\Lambda^T(\mathcal{B}_{\text{col}}(\mathbf{Z}) - \mathbf{V}_1)) \\ + \frac{\mu_1}{2}||\mathbf{V}_1 - \mathcal{B}_{\text{col}}(\mathbf{Z})||_F^2. \quad (15)$$

Referring to [48], the update of $\mathbf{V}_1$ is determined by the proximal mapping

$$\mathbf{V_1} = \arg\min_{\mathbf{V_1}} \frac{1}{2}\left\{[\mathbf{V}_1 - \mathcal{B}_{\text{col}}(\mathbf{Z})]_{\text{col},2,2} + \frac{\gamma}{\mu_1}[\mathbf{V}_1]_{\text{col},2}\right\} \\ = \text{prox}_{\sigma_{\text{col}}||\cdot||}\left(\mathcal{B}_{\text{col}}(\mathbf{Z}) - \frac{1}{\mu_1}\Lambda\right) \quad (16)$$

where $\sigma_{\text{col}} = [(\gamma/||\mathbf{V}_{1,1}||_2), \ldots, (\gamma/||\mathbf{V}_{1,|E|}||_2)]$. For a matrix $\mathbf{M} \in \mathcal{R}^{p\times q}$, we define $[\mathbf{M}]_{\text{col},2,2} = [||\mathbf{M}_{.1}||_2^2, \ldots, ||\mathbf{M}_{.q}||_2^2]$. Note that $\text{prox}_{\sigma_{\text{col}}||\cdot||}(\mathbf{M})$ does the proximal operation on each column of $\mathbf{M}$.

*Step 3 (Fix Others, Update $\mathbf{V}_2$):* To update $\mathbf{V}_2$, we need to optimize the following function:

$$f(\mathbf{V_2}) = \gamma\mathbf{w}^T[\mathbf{V}_2]_{\text{row},2} + \text{tr}(\Psi^T(\mathcal{B}_{\text{row}}(\mathbf{Z}) - \mathbf{V}_2)) \\ + \frac{\mu_2}{2}||\mathbf{V}_2 - \mathcal{B}_{\text{row}}(\mathbf{Z})||_F^2. \quad (17)$$

Similar with the update of $\mathbf{V}_1$, the update of $\mathbf{V}_2$ is also determined by the proximal mapping

$$\mathbf{V_2} = \arg\min_{\mathbf{V_2}} \frac{1}{2}\left\{[\mathbf{V}_2 - \mathcal{B}_{\text{row}}(\mathbf{Z})]_{\text{row},2,2} + \frac{\gamma}{\mu_2}[\mathbf{V}_2]_{\text{row},2}\right\} \\ = \text{prox}_{\sigma_{\text{row}}||\cdot||}\left(\mathcal{B}_{\text{row}}(\mathbf{Z}) - \frac{1}{\mu_2}\Psi\right) \quad (18)$$

where $\sigma_{\text{row}} = [(\gamma/||\mathbf{V}_{2,1.}||_2), \ldots, (\gamma/||\mathbf{V}_{2|E|.}||_2)]^T$. For a matrix $\mathbf{M} \in \mathcal{R}^{p\times q}$, we define $[\mathbf{M}]_{\text{row},2,2} = [||\mathbf{M}_{1.}||_2^2, \ldots, ||\mathbf{M}_{p.}||_2^2]^T$. Note that $\text{prox}_{\sigma_{\text{row}}||\cdot||}(\mathbf{M})$ does the proximal operation on each row of $\mathbf{M}$.

*Step 4 (Fix Others, Update $\Lambda$):* To update the Lagrange multiplier matrix $\Lambda$, we use

$$\Lambda = \Lambda + \mu_1(\mathbf{V}_1 - \mathcal{B}_{\text{col}}(\mathbf{Z})). \quad (19)$$

*Step 5 (Fix Others, Update $\Psi$):* To update the Lagrange multiplier matrix $\Psi$, we use the following formula:

$$\Psi = \Psi + \mu_2(\mathbf{V}_2 - \mathcal{B}_{\text{row}}(\mathbf{Z})). \quad (20)$$

The whole procedure to obtain the solution of (10) is given in Algorithm 1.

---

**Algorithm 1** ABDR Solver

---

**Inputs:** $\mathbf{X} \in \mathcal{R}^{d\times n}$, $\mathcal{Q}$, $\mathbf{w}$, $\gamma$, MaxIter.
1: **Initialization:** $k = 0$, $\mathbf{Z}^k$, $\mathbf{V}_1^k$, $\mathbf{V}_2^k$, $\Lambda^k$, $\Psi^k$.
2: **for** each $k \in \{1, 2, 3, \ldots,$ MaxIter$\}$ **do**
3:     Update $\mathbf{Z}^k$ by Eq.(12);
4:     Update $\mathbf{V}_1^k$ by Eq.(16);
5:     Update $\mathbf{V}_2^k$ by Eq.(18);
6:     Update $\Lambda^k$ by Eq.(19);
7:     Update $\Psi^k$ by Eq.(20);
8:     **if** converged **then**
9:         Break
10:    **end if**
11: **end for**
12: **return** $\mathbf{Z}$.

---

*3) Complexity Analysis:* Now, we give the computational complexity of GADMM for optimizing our ABDR. In our implementation, each update of $\mathbf{Z}$ involves $(\mathbf{X}^T\mathbf{X} + \alpha\mathbf{I})^{-1}$, which needs the computational complexity of $\mathcal{O}(n^3)$. But note that we need to compute $(\mathbf{X}^T\mathbf{X} + \alpha\mathbf{I})^{-1}$ only once. In each iteration, the updates of $\mathbf{Z}$, $\mathbf{V}_1$, $\mathbf{V}_2$, $\Lambda$, $\Psi$ need complexity of $\mathcal{O}(n^2d)$, $\mathcal{O}(n^2)$, $\mathcal{O}(n^2)$, $\mathcal{O}(C)$, and $\mathcal{O}(C)$, respectively, where $C$ is a constant. So the computational complexity of ABDR is $\mathcal{O}(T_1n^2d + n^3)$, where $T_1$ is the total number of iterations until Algorithm 1 converges. For BDR, the computation is dominated by the updates of $\mathbf{B}$ and $\mathbf{Z}$, which needs to compute the eigenvectors of the matrix of size $n \times n$ ($\mathcal{O}(n^3)$). So the computational complexity of ABDR is $\mathcal{O}(T_2n^3)$, where $T_2$ is the total number of iterations until BDR converges. Moreover, from Fig. 12(a) and (b), we can observe that $T_1$ is very small (about 20). In summary, our proposed method ABDR is more efficient than BDR.

*4) Convergence Analysis:* Although the convergence of inexact ALM has been strictly proven when the number of blocks is at most two [44], its convergence properties for minimizing the objective function with more than two primal variables subjected to linear constraints remains generally unclear [45]. The equivalent form (9) of objective function of ABDR has three primal variables and is not smooth, at the first glance, it seems difficult to prove its convergence in theory. However, due to the uniqueness of our objective function and as indicated in [33], two conditions are sufficient for our Algorithm 1 to converge: 1) the feature matrix $\mathbf{X}$ is of full column rank and 2) the optimality gap generated in each iteration is monotonically decreasing. The first condition can be easily met by factorizing $\mathbf{Z}$ into $\mathbf{PZ}$, where $\mathbf{P}$ can be

TABLE II
STATISTICS OF THE BENCHMARK DATASETS

| Dataset | # instances | # features | # clusters |
|---|---|---|---|
| Hopkins 155* | - | - | - |
| Extended YaleB | 2414 | 2016 | 38 |
| PIE | 2856 | 1024 | 68 |
| MNIST | 6996 | 784 | 10 |

*Because Hopkins 155 database covers 155 datasets with different number of features and dimensions, we do not show its statistic information in Table II but describe it in detail in Section IV.B.

computed in advance by orthogonalizing the columns of $\mathbf{X}$. The convexity of the Lagrangian function (10) can guarantee the second condition to some extent [46]. In conclusion, the GADMM for solving our ABDR converges to the global solution and fast as demonstrated in Fig. 12. **Theorem** 2 gives the theoretical guarantee for the convergence of Algorithm 1. The detailed proof of **Theorem** 2 can be found in the supplementary material.

*Theorem 2:* **Algorithm 1** converges to the global minimizer of (5).

### C. Subspace Clustering Algorithm

We describe the process of ABDR for subspace clustering. For a given data matrix $\mathbf{X}$, we first obtain the coefficient matrix $\mathbf{Z}$ by solving the optimization problem (5) using Algorithm 1 and construct the affinity matrix $\mathbf{W} = (|\mathbf{Z}| + |\mathbf{Z}^T|)/2$. Next, we apply spectral clustering [49] on $\mathbf{W}$ to determine the final partion/segmentation.

Different from BDR, our proposed ABDR does not need to prefix the number of subspaces/blocks when computing the coefficient matrix. Moreover, we also empirically observe that the number of diagonal blocks of coefficient matrix $\mathbf{Z}$ obtained by ABDR exactly matches to the number of subspaces, thus providing a reference for spectral clustering when the number of subspaces is unknown.

## IV. EXPERIMENTS

For fair comparison with existing SOTAs, especially BDR, we follow the protocols of BDR without considering deep learning based subspace clustering methods and select the following methods for comparison: $K$-means [50] (a classic clustering method), ratio cut (Rcut) [51], normalized cut (Ncut) [52] (two representative locality-inducing spectral-type subspace clustering methods), SSC, LRR, BDR (three SOTA globality-inducing spectral-type subspace clustering methods), structured block diagonal representation (SBDR) [53] (an extension of BDR which incorporates the cluster assignment into the learning of representation coefficient matrix) as the compared methods. We test these methods and ABDR on three synthetic examples and four benchmark real datasets for various subspace clustering tasks: Hopkins155 database for motion segmentation [27], [31], [33], Extended YaleB database [27], [31], [33], [57], and pose illumination and expression (PIE) database [58] for face clustering, Mixed National Institute of Standards and Technology (MNIST) database for handwritten digit clustering [27]. The statistic information of these benchmark datasets is shown in Table II.

Note we consider the settings and the ways to construct data matrix $\mathbf{X}$ adopted in BDR [27].

Similar with BDR, we utilize the clustering error defined as follows to evaluate the clustering performance:

$$\text{clustering error} = 1 - \frac{1}{n}\sum_{i=1}^{n}\delta(p_i - \text{map}(q_i)) \quad (21)$$

where $p_i$ and $q_i$ are the predictive label and ground truth label of the $i$th data point, respectively. $\delta(x, y) = 1$ if $x = y$, $\delta(x, y) = 0$ otherwise. map($\cdot$) is the optimal mapping function that permutes predictive labels to maximally match with the ground truth labels. Note that the lower the value of clustering error is, the better the clustering performance is.

For each method, we tune the parameter(s) in a wide range and utilize the one(s) which obtains the best result in most cases for each dataset. Because $k$-means is sensitive to initialization, we run $k$-means 20 times and report the best result. For some methods over certain datasets, we use the parameters given in their articles or codes. Note we find that the postprocess influences the final clustering result at certain extent, so we perform the optimal postprocess for each obtained $\mathbf{Z}$ ($\mathbf{B}$). This explains the fact that the clustering errors of certain compared methods for some datasets in this article are lower than that shown in [27].

Next, Section IV-A gives the synthetic illustration, Sections IV-B–IV-D validate the effectiveness of the proposed method ABDR in real scenarios. Section IV-E discusses the behavior of hyperparameter in the objective function of ABDR on real datasets. Section IV-F verifies the necessity of two fused terms in ABDR.

### A. Synthetic Illustration

*Example 1:* Here, we give a simple example to illustrate the effectiveness of ABDR. As Fig. 1(a) shows, we sample 30 data points from two $1D$ subspaces corresponding to $y = x$ and $y = -x$, respectively, in $\mathcal{R}^2$ to construct the data matrix $\mathbf{X} = \{\mathbf{X_1}, \mathbf{X_2}\}$, where $\mathbf{X_1}$ includes 20 data points and $\mathbf{X_2}$ includes 10 data points, respectively. The coefficient matrices obtained by BDR and ABDR are shown in Figs. 2(a) and (b) and 4(a), respectively. As Fig. 2(a) and (b) shows, even if the correct number of subspaces is given, BDR still obtains both coefficient matrices with three diagonal blocks, which is different from the ground-truth. This illustrates that it is difficult for BDR to obtain the exact coefficient matrix that can reflect true global subspace structures on this dataset. Besides, we exploit KNN with number of neighbors $K = 5$ and $K = 10$ to construct the coefficient matrices for this example and show them in Fig. 3. From Fig. 3, we can see that KNN cannot construct a coefficient matrix to reflect the subspace structure of this dataset. Differently, from Fig. 4(a), we can find that ABDR adaptively acquires the exact BDR matrix where the number of blocks matches to the ground-truth even without predefinition, performing better than BDR. Next, we further validate the effect of given number of subspaces on BDR and show the obtained $\mathbf{B}$ and $\mathbf{Z}$ by BDR in Fig. 2(a)–(h). From them, we can observe that worse representation matrices are obtained by BDR over the incorrect number of subspaces. Compared with BDR, ABDR can adaptively construct the more exact block diagonal matrix without prefixing the number of subspaces, thus not suffering from this problem. Besides, Fig. 4 shows that ABDR can still obtain the coefficient matrices with exact block diagonal structure over several hyperparameters with large differences, this illustrates the low sensitivity of ABDR to hyperparameter to a certain extent.

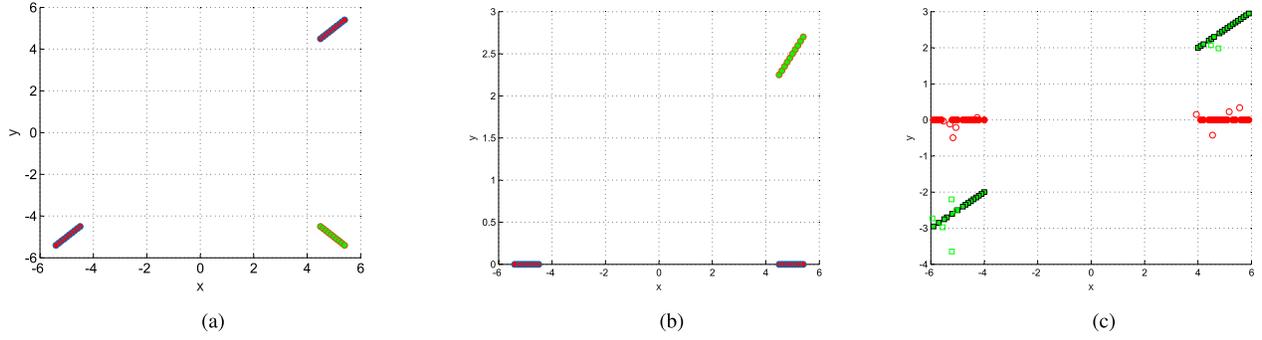

Fig. 1. Three synthetic datasets. (a) 30 data points sampled two $1D$ subspaces including $y = x$ and $y = -x$, where 20 data points are sampled from the subspace $y = (1/2)x$ and 10 data points are sampled from the subspace $y = -x$, respectively. (b) 30 data points sampled from two $1D$ subspaces including $y = 0$ and $y = x$, where 20 data points are sampled from the subspace $y = 0$ and 10 data points are sampled from the subspace $y = x$, respectively. (c) 80 data points sampled from two $1D$ subspaces $y = 0$ and $y = (1/2)x$ with Gaussian noisy rate $\gamma = 0.2$, where 40 data points are sampled from the subspace $y = 0$ and 40 data points are sampled from the subspace $y = (1/2)x$, respectively. (a) Example 1. (b) Example 2. (c) Example 3.

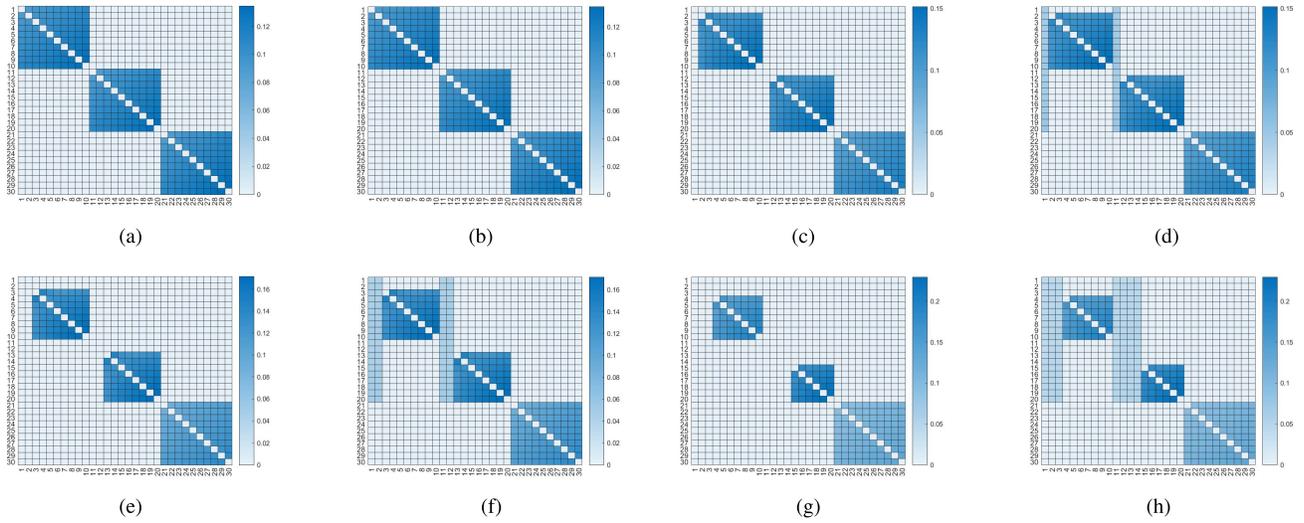

Fig. 2. **B** and **Z** obtained by BDR over different number of subspaces given on Example 1. (a) BDR_B, given $k = 2$. (b) BDR_Z, given $k = 2$. (c) BDR_B, given $k = 5$. (d) BDR_Z, given $k = 5$. (e) BDR_B, given $k = 7$. (f) BDR_Z, given $k = 7$. (g) BDR_B, given $k = 10$. (h) BDR_Z, given $k = 10$.

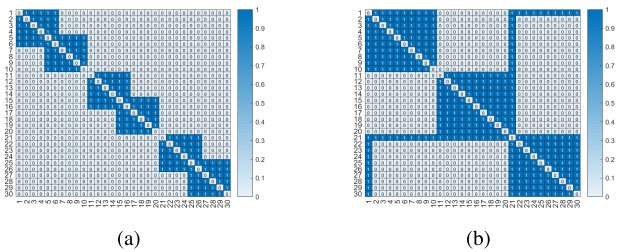

Fig. 3. **Z** constructed by KNN on Example 1. (a) $K = 5$. (b) $K = 10$.

*Example 2:* Here, we give another synthetic example to illustrate the effectiveness of ABDR. We generate the data matrix $\mathbf{X} = \{\mathbf{X_1}, \mathbf{X_2}\}$ with its columns drawn from 2 $1D$ subspaces without noise, where the two subspaces correspond to $y = 0$ and $y = (1/2)x$, respectively. We randomly choose 20 data points from the $1D$ subspace of $y = 0$ and 10 data points from the $1D$ subspace of $y = (1/2)x$ to construct $\mathbf{X}$ and show it in Fig. 1(b). We perform both ABDR and BDR on this dataset and show the resulting coefficient matrices in Figs. 5 and 6, respectively. From both figures, we can find that BDR still obtains the block diagonal coefficient matrix with three blocks, not reflecting the true global subspace structures of the dataset. Different from BDR, ABDR determines the exact BDR matrix where the number of blocks matches with the true number of subspaces, even without predefinition. This further demonstrates the superiority of ABDR in reflecting the subspace structure of data points.

*Example 3:* Here, we illustrate the effectiveness and robustness of ABDR on the noisy data. We first generate the data matrix $\mathbf{X} = \{\mathbf{X_1}, \mathbf{X_2}\}$ where its columns are drawn from 2 $1D$ subspaces without noise. The two subspaces correspond to $y = 0$ and $y = (1/2)x$, respectively. We randomly choose 40 data points from the $1D$ subspace of $y = 0$ and 40 data points from the $1D$ subspace of $y = (1/2)x$. Then we randomly add Gaussian noise with $\mu = [0\ 0; 0\ 0]$ and $\Sigma = [0.1\ 0; 0\ 0.1]$ ($\mu$ and $\Sigma$ are mean matrix and standard error matrix, respectively) to the 80 data points with noisy rate $r = 0.2$ to generate the noisy matrix **X** and show it in Fig. 1(c). The coefficient matrices obtained by the ABDR and BDR are shown in Figs. 7 and 8, respectively. We can see that ABDR obtains a desirable block diagonal coefficient matrix where the number of blocks exactly matches to the number of subspaces, illustrating the robustness of ABDR to noisy data points and the ability of ABDR to reflect the true global subspace structure of the dataset. On the other hand, BDR obtains the block diagonal coefficient matrices with four blocks, not matching with the true one. This example further demonstrates better performance of ABDR than that of BDR on noisy observations.

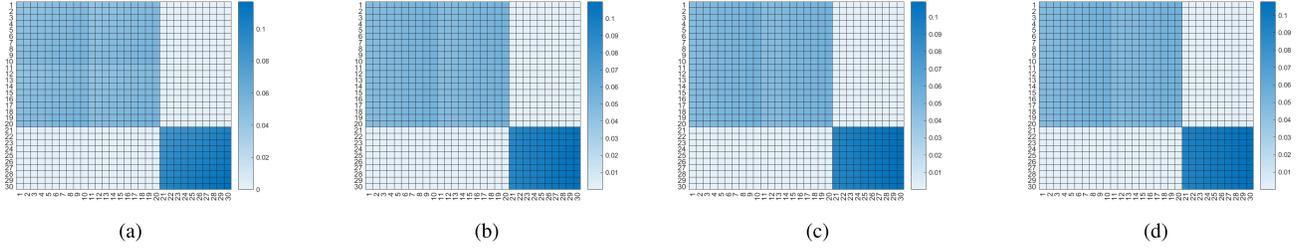

Fig. 4. **Z** obtained by ABDR over different hyperparameters on Example 1. (a) $\gamma = 0.001$. (b) $\gamma = 1$. (c) $\gamma = 10$. (d) $\gamma = 1000$.

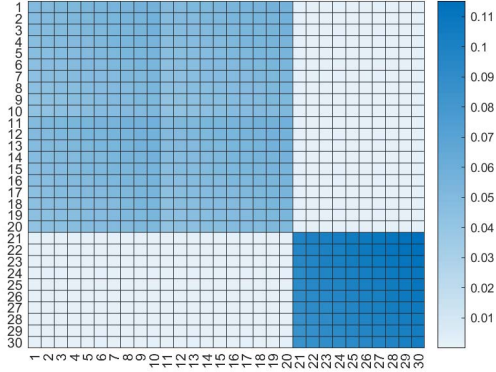

Fig. 5. **Z** obtained by ABDR on Example 2.

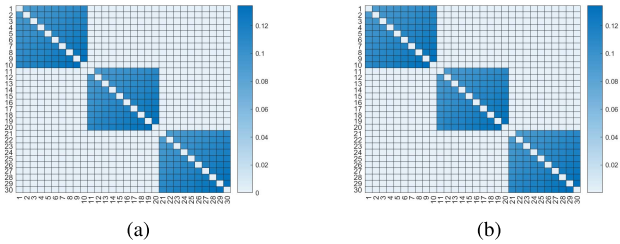

Fig. 6. **B** and **Z** obtained by BDR on Example 2. (a) **B**. (b) **Z**.

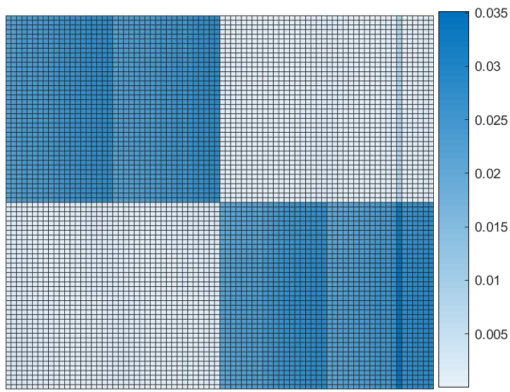

Fig. 7. **Z** obtained by ABDR on Example 3.

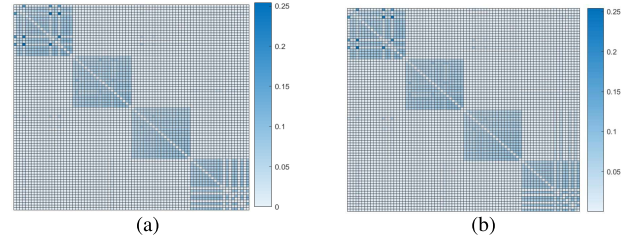

Fig. 8. **B** and **Z** obtained by BDR on Example 3. (a) **B**. (b) **Z**.

### B. Motion Segmentation

Here, we illustrate the effectiveness of ABDR by considering the application of subspace clustering on motion segmentation, which refers to the problem of segmenting the video sequences into multiple spatiotemporal regions where each region corresponds to a rigid-body motion in the scene. As [30] showed, the coordinates of the points in trajectories of one moving object reside in a 3-D subspace. Therefore, the problem of motion segmentation can be solved by subspace clustering. In this section, following BDR, we choose widely used Hopkins155 database [54]. It consists of 155 video sequences, where 120 video sequences have two motions and 35 video sequences have three motions. Each sequence is a whole dataset so there are total 155 subspace clustering tasks.

Similar with the setting in [27], we consider using two forms to construct the data matrix **X** for each video sequence: 1) directly use the original $2F$-dimensional feature trajectories, where $F$ is the number of frame of the sequence and 2)

TABLE III
CLUSTERING ERRORS (%) OF DIFFERENT METHODS ON THE HOPKINS 155 DATABASE WITH THE $2F$-DIMENSIONAL DATA POINTS

| method | K-means | Rcut | Ncut | SSC | LRR | BDR-B | BDR-Z | SBDR | Ours |
|---|---|---|---|---|---|---|---|---|---|
| 2 motions | | | | | | | | | |
| mean | 19.55 | 15.98 | 16.94 | 0.63 | 0.85 | **0.54** | 0.61 | 1.95 | 0.75 |
| median | 17.92 | 8.96 | 11.19 | 0.00 | 0.00 | 0.00 | 0.00 | 0.00 | 0.00 |
| 3 motions | | | | | | | | | |
| mean | 26.02 | 24.97 | 25.00 | 2.95 | 3.07 | **0.36** | 0.39 | 1.59 | 2.39 |
| median | 20.48 | 20.73 | 26.42 | 0.21 | 0.56 | 0.00 | 0.00 | 0.00 | 0.71 |
| All motions | | | | | | | | | |
| mean | 21.02 | 18.01 | 18.76 | 1.16 | 1.35 | **0.49** | 0.56 | 3.19 | 1.12 |
| median | 18.99 | 15.55 | 16.98 | 0.00 | 0.00 | 0.00 | 0.00 | 0.19 | 0.00 |

- The lowest index value is highlighted in **bold**, the second lowest is underlined. The lower the value of the index is, the better the clustering performance is. The same meaning for the following tables.

TABLE IV
CLUSTERING ERRORS (%) OF DIFFERENT METHODS ON THE HOPKINS 155 DATABASE WITH THE $4k$-DIMENSIONAL DATA POINTS BY USING PCA

| method | K-means | Rcut | Ncut | SSC | LRR | BDR-B | BDR-Z | SBDR | Ours |
|---|---|---|---|---|---|---|---|---|---|
| 2 motions | | | | | | | | | |
| mean | 19.56 | 15.98 | 16.94 | 0.68 | 1.36 | 0.39 | **0.38** | 1.93 | 0.98 |
| median | 17.92 | 8.96 | 11.19 | 0.00 | 0.00 | 0.00 | 0.00 | 0.00 | 0.00 |
| 3 motions | | | | | | | | | |
| mean | 26.02 | 24.97 | 25.08 | 2.97 | 2.95 | **0.38** | **0.38** | 1.48 | 2.36 |
| median | 20.48 | 20.73 | 26.60 | 0.20 | 0.56 | 0.00 | 0.00 | 0.00 | 0.71 |
| All motions | | | | | | | | | |
| mean | 21.02 | 18.01 | 18.78 | 1.20 | 1.72 | 0.39 | **0.38** | 3.45 | 1.29 |
| median | 18.99 | 15.55 | 16.98 | 0.00 | 0.00 | 0.00 | 0.00 | 0.19 | 0.00 |

use principal component analysis (PCA) to project the whole data matrix into the $4k$-dimensional subspace, where $k$ is the number of subspaces. For all of the spectral-based methods, as the same setting in [31], we define the affinity matrix $\mathbf{W} = (|\mathbf{Z}| + |\mathbf{Z}^T|)/2$, where $\mathbf{Z}$ is the obtained coefficient matrix by different methods. Hopkins155 database consists of 120 video sequences of 2 motions and 35 video sequences of 3 motions, so we report the segmentation results of 2 motions, 3 motions and all video sequences over two settings in Tables III and IV, respectively.

Based on the results in Tables III and IV, we have the following observations and conclusions.

1) We report the means of clustering errors of many video sequences, among which the values of clustering errors of BDR almost approximate zero. Moreover, the other compared methods such as SSC already perform very well. So it is too challenging to obtain considerable improvements on this motion segmentation. Even so, our ABDR just performs slightly lower than BDR while outperforms other compared methods on the setting of $2F$ with clustering accuracy approximating to 99% (100%–1.12%). Although ABDR performs slightly lower than BDR and SSC on the setting of $4k$, it still obtains clustering accuracy about 99% (100%–1.29%). Note both SSC has higher computational cost than ABDR. Besides, Tables III and IV report the means of clustering errors of many (155) subspace clustering tasks, ABDR performs well on this motion segmentation with clustering accuracy 99% on both settings. Furthermore, ABDR only has one hyperparameter while BDR needs to adjust two hyperparameters, making its model determination more complicated than ABDR. Besides, the nonconvexity of BDR makes it easily get into local minima. Although SBDR incorporates the cluster assignment into the learning of coefficient matrix while ABDR learns the coefficient matrix without the help of the cluster assignment, ABDR perform better than SBDR on both settings. So in summary, ABDR is a good choice in real applications.

### C. Face Clustering

Face clustering aims to partition face images into clusters according to their respective subjects. Under the Lambertian assumption, the face images corresponding to a subject with a certain pose and varying illumination approximately reside in a subspace of dimension 9 [55]. Therefore, a set of face images of multiple subjects approximately reside in a union of 9-D linear subspaces. Naturally, the problem of face clustering can be solved by subspace clustering method. Here, we validate the effectiveness of ABDR on two widely used face datasets: Extended YaleB [56] and PIE [58].

*1) Extended YaleB:* Extended YaleB dataset consists of 2414 face images of 38 subjects including nine poses and 64 illumination conditions, where each subject covers 64 images. The original size of each face image is $192 \times 168$. To reduce the computational and memory cost, as what was done in [27], we downsample each image to the size of $48 \times 32$ and vectorize it to a vector of length 2016 as a data point. Meanwhile, we normalize each data point with a unit length. Next, we construct the data matrix $\mathbf{X}$ by the subset with $k \in \{2, 3, 5, 8, 10\}$ subjects from the Extended YaleB database. For each $k$, we randomly choose $k$ subjects from the total 38 subjects and each subset consists of $64k$ face images. For each $k$, we perform 20 trails and report the mean, median and standard deviation of clustering error.

TABLE V
CLUSTERING ERRORS (%) OF DIFFERENT METHODS ON THE EXTENDED YALEB DATABASE

| Algorithm | 2 subjects | | | 3 subjects | | | 5 subjects | | | 8 subjects | | | 10 subjects | | |
|---|---|---|---|---|---|---|---|---|---|---|---|---|---|---|---|
| | mean | median | std | mean | median | std | mean | median | std | mean | median | std | mean | median | std |
| K-means | 48.24 | 48.44 | 1.54 | 64.11 | 64.06 | 1.18 | 77.08 | 77.34 | 1.07 | 84.25 | 84.18 | 0.75 | 86.24 | 86.25 | 0.47 |
| Rcut | 46.84 | 47.66 | 2.69 | 54.01 | 51.56 | 9.76 | 61.09 | 60.16 | 3.71 | 65.74 | 67.48 | 5.50 | 67.77 | 68.52 | 3.31 |
| Ncut | 45.74 | 46.48 | 3.32 | 54.11 | 54.95 | 9.73 | 59.23 | 58.75 | 3.94 | 63.48 | 65.14 | 6.43 | 66.16 | 67.03 | 4.52 |
| SSC | 0.74 | 0.00 | 1.61 | 3.02 | 0.52 | 7.17 | 4.14 | 1.41 | 4.10 | 7.56 | 8.30 | 5.64 | 9.38 | 9.53 | 6.88 |
| LRR | 3.01 | 1.17 | 5.76 | 3.59 | 2.60 | 3.94 | 5.08 | 4.06 | 3.97 | 6.57 | 4.59 | 5.40 | 8.34 | 5.94 | 5.71 |
| BDR-B | 1.80 | 0.00 | 6.61 | 2.42 | 0.52 | 7.91 | 3.09 | 2.03 | 4.08 | **2.39** | 2.34 | 0.81 | 2.74 | 2.50 | 1.08 |
| BDR-Z | 0.90 | 0.00 | 3.00 | **0.57** | 0.52 | 0.71 | **1.84** | 1.41 | 1.44 | 2.79 | 2.83 | 1.34 | 2.60 | 2.50 | 1.22 |
| SBDR | **0.00** | 0.00 | 0.00 | 1.88 | 1.88 | 0.28 | 3.45 | 3.45 | 0.00 | 3.92 | 3.92 | 0.00 | **0.79** | 0.79 | 0.00 |
| Ours | 0.74 | 0.00 | 1.47 | 2.27 | 0.78 | 5.41 | 3.20 | 2.03 | 2.76 | 2.79 | 2.34 | 2.70 | 4.16 | 2.97 | 3.53 |

The clustering errors of different subspace clustering methods on the Extended YaleB database are reported in Table V. From it, we can see that three direct learning methods consistently perform better than the indirect learning methods with a large margin. This sufficiently illustrate the superiority of direct learning methodology to indirect learning one. Besides, we can see that ABDR outperforms BDR on two subjects of Extended YaleB. This demonstrates that ABDR may especially performs well in small dataset, note the small dataset commonly exists in real applications. Actually, from Table V, we can see that ABDR or BDR has performed very well on the this dataset with different number of subjects, so it is too challenging to obtain considerable improvements. Even so, ABDR performs well with relatively small clustering errors over other cases. Moreover, although the face clustering task becomes more challenging as the number of subjects increases, the improvements of ABDR become more significant as the number of subjects increases compared with other existing representative subspace clustering methods such as LRR and SSC. Therefore, our proposed method ABDR is effective to deal with challenging face clustering task. Here, note that although both SBDR and BDR perform slightly better than ABDR on some subsets, both of them correspond to a nonconvex objectives with more than one hyperparameters and need to prefix the number of subspaces, leading to its sensitivity to initialization, local minimum solution and the difficulty of real implementation. In contrast, ABDR corresponds to a convex objective with only one parameter, so ABDR is a good choice for subspace clustering in real applications.

We provide the average computational time of each method as a function of the number of subjects on Extended YaleB in Fig. 9. From Fig. 9, we can see that both ABDR and BDR have relatively low running time and work much efficient than SSC, LRR and SBDR, this further validates the low computational cost of ABDR.

*2) PIE:* PIE database includes 2856 face images of 68 people, each of which has $32 \times 32$ pixels. We vectorize each face image to a 1024 data vector and normalize it to a vector with unit length. Each people has 42 face images under different light and illumination conditions. Similar to the way of constructing the data matrix **X** for Extended YaleB, we construct the data matrix **X** by the subset which consists of $k \in \{10, 20, 40, 50, 68\}$ subjects. For each $k$, we randomly choose $k$ people from the total 68 people and use the $42k$ data points to construct the data matrix **X**. For each $k$, we perform

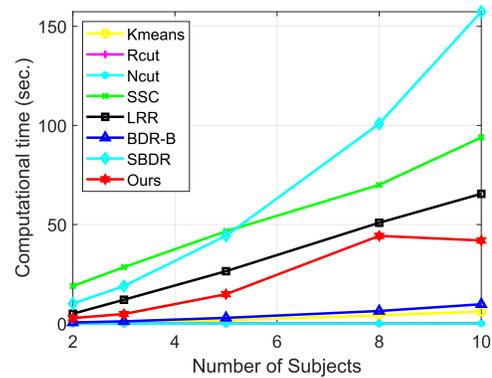

Fig. 9. Average computational time with respect to the number of subjects on Extended YaleB.

the clustering methods for 20 trials and report the mean, median, and standard deviation.

The clustering errors by different subspace clustering methods on PIE database are shown in Table VI. Table VI shows that our ABDR outperforms in most of cases except for the dataset with 68 subjects. Although SBDR performs the best on 68 subjects with clustering error of 0, our proposed ABDR achieves clustering error of 0.0039 which almost approximates 0. Note it is relatively unfair to compare ABDR with SBDR because SBDR incorporates the cluster assignment into the learning of coefficient learning while ABDR learns the coefficient matrix without the help of cluster assignment. Even so, ABDR determines the similar performance with SBDR. Although the clustering task especially becomes more challenging as the number of subjects increases, ABDR still improves 2.81% in clustering error than that of BDR on the 68 subjects. Meanwhile, Table VI reports the average result in clustering error, therefore the improvement obtained by ABDR is significant. This experiment demonstrates the effectiveness of ABDR for the face clustering task on the PIE database. Moreover, ABDR has single hyperparameter to adjust, leading to easy model determination. Therefore, ABDR is a good choice for real subspace clustering tasks.

### D. Handwritten Digit Clustering

*MNIST:* The images of handwritten digits reside in the subspaces of dimension 12 [59], so the problem of clustering images of handwritten digits can be solved by subspace clustering method. Here, we choose MNIST database to validate the effectiveness of ABDR, which consists of gray scale images of ten subjects, i.e., 0–9 handwritten digits. Each gray image of MNIST database has $28 \times 28$ pixels and we vectorize each image into a data vector of length 784. Each data vector is normalized as one with unit length. We consider the problem of clustering $k$ subjects, where $k$ varies from two to ten. For each $k$, we perform each subspace clustering method for 20 trials and report the average of clustering error. Note that for each $k$ and each trial, we randomly choose $k$ subjects from the total ten subjects and each subject consists of 100 samples. That is, we construct the data matrix **X** for each $k$ and each trial of size $784 \times 100 \ k$.

TABLE VI
CLUSTERING ERRORS (%) OF DIFFERENT METHODS ON THE PIE DATABASE

| Algorithm | 10 subjects | | | 20 subjects | | | 40 subjects | | | 50 subjects | | | 68 subjects | | |
|---|---|---|---|---|---|---|---|---|---|---|---|---|---|---|---|
| | mean | median | std | mean | median | std | mean | median | std | mean | median | std | mean | median | std |
| K-means | 70.50 | 70.60 | 3.43 | 73.51 | 73.51 | 1.70 | 74.50 | 74.35 | 1.21 | 75.92 | 75.79 | 0.83 | 76.33 | 76.30 | 1.10 |
| Rcut | 17.12 | 16.90 | 5.56 | 18.94 | 19.11 | 4.37 | 26.75 | 26.55 | 2.61 | 29.43 | 29.12 | 2.46 | 32.74 | 32.77 | 1.73 |
| Ncut | 14.94 | 14.64 | 5.22 | 14.69 | 15.48 | 5.21 | 19.51 | 19.52 | 3.33 | 22.42 | 22.52 | 2.65 | 25.24 | 25.40 | 0.57 |
| SSC | 8.27 | 9.52 | 7.13 | 4.73 | 4.76 | 4.11 | 3.59 | 3.57 | 1.92 | 3.68 | 3.81 | 1.72 | 3.47 | 3.78 | 1.30 |
| LRR | 1.71 | 0.00 | 3.82 | 0.35 | 0.00 | 1.13 | 0.97 | 0.00 | 1.52 | 0.92 | 0.00 | 1.29 | 2.74 | 2.50 | 1.08 |
| BDR-B | **0.00** | 0.00 | 0.00 | **0.00** | 0.00 | 0.00 | **0.00** | 0.00 | 0.00 | **0.00** | 0.00 | 0.00 | 6.44 | 6.44 | 0.00 |
| BDR-Z | **0.00** | 0.00 | 0.00 | **0.00** | 0.00 | 0.00 | 0.16 | 0.00 | 0.72 | 1.20 | 0.00 | 1.36 | 3.20 | 3.78 | 1.15 |
| SBDR | **0.00** | 0.00 | 0.00 | **0.00** | 0.00 | 0.00 | **0.00** | 0.00 | 0.00 | **0.00** | 0.00 | 0.00 | **0.00** | 0.00 | 0.80 |
| Ours | **0.00** | 0.00 | 0.00 | **0.00** | 0.00 | 0.00 | **0.00** | 0.00 | 0.00 | **0.00** | 0.00 | 0.00 | 0.39 | 0.00 | 0.80 |

TABLE VII
CLUSTERING ERRORS (%) OF DIFFERENT METHODS ON THE MNIST DATABASE

| Algorithm | 2 subjects | | | 3 subjects | | | 4 subjects | | | 5 subjects | | | 6 subjects | | |
|---|---|---|---|---|---|---|---|---|---|---|---|---|---|---|---|
| | mean | median | std | mean | median | std | mean | median | std | mean | median | std | mean | median | std |
| K-means | 6.48 | 4.25 | 8.22 | 24.15 | 24.33 | 15.78 | 29.48 | 31.25 | 9.84 | 35.05 | 35.40 | 10.52 | 40.76 | 41.17 | 6.39 |
| Rcut | 5.10 | 1.75 | 7.55 | 19.18 | 17.00 | 13.80 | 23.06 | 21.63 | 11.93 | 31.30 | 32.00 | 8.11 | 38.25 | 38.58 | 8.12 |
| Ncut | 4.10 | 2.25 | 5.58 | 15.30 | 9.67 | 12.91 | 23.01 | 23.75 | 12.03 | 28.86 | 31.10 | 9.82 | 36.84 | 38.00 | 7.42 |
| SSC | 4.20 | 1.50 | 7.17 | 14.67 | 13.00 | 10.91 | 21.19 | 18.00 | 10.45 | 24.81 | 27.20 | 8.05 | 34.58 | 33.42 | 6.45 |
| LRR | 3.05 | 1.50 | 4.27 | 14.57 | 9.00 | 13.47 | 20.14 | 17.00 | 10.79 | 27.23 | 30.70 | 9.10 | 34.32 | 35.83 | 6.20 |
| BDR-B | 9.65 | 1.25 | 17.39 | 13.08 | 8.00 | 11.85 | 19.16 | 16.63 | 10.26 | 26.50 | 29.40 | 10.08 | 35.68 | 38.00 | 7.17 |
| BDR-Z | 4.93 | 1.50 | 11.41 | 11.93 | 5.83 | 12.56 | 19.33 | 16.13 | 11.14 | 24.91 | 27.90 | 10.22 | 34.74 | 35.42 | 6.17 |
| SBDR | 4.58 | 0.75 | 10.92 | 17.17 | 16.50 | 14.38 | 19.66 | 19.88 | 10.70 | 24.12 | 27.80 | 9.79 | 33.23 | 35.08 | 6.14 |
| Ours | **1.98** | 1.00 | 3.35 | **9.77** | 3.33 | 12.10 | **16.85** | 14.50 | 11.09 | **23.92** | 28.80 | 10.44 | **32.25** | 32.58 | 5.41 |

TABLE VIII
CLUSTERING ERRORS (%) OF DIFFERENT METHODS ON THE MNIST DATABASE

| Algorithm | 7 subjects | | | 8 subjects | | | 9 subjects | | | 10 subjects | | |
|---|---|---|---|---|---|---|---|---|---|---|---|---|
| | mean | median | std | mean | median | std | mean | median | std | mean | median | std |
| K-means | 38.71 | 40.29 | 5.77 | 42.35 | 42.44 | 4.49 | 45.03 | 45.89 | 3.74 | 47.01 | 45.95 | 3.38 |
| Rcut | 36.28 | 35.36 | 6.74 | 41.15 | 42.69 | 7.33 | 42.03 | 42.44 | 6.07 | 42.09 | 41.75 | 2.77 |
| Ncut | 33.86 | 33.36 | 6.45 | 37.76 | 39.25 | 7.55 | 40.26 | 40.83 | 6.41 | 40.56 | 40.20 | 2.60 |
| SSC | 32.02 | 32.57 | 5.65 | 39.56 | 39.19 | 6.43 | 40.11 | 40.50 | 4.52 | 41.98 | 42.00 | 2.57 |
| LRR | 31.83 | 30.86 | 5.38 | 36.22 | 36.19 | 4.78 | 40.44 | 39.83 | 4.44 | 41.60 | 41.40 | 2.73 |
| BDR-B | 31.76 | 30.50 | 5.79 | 35.94 | 36.63 | 5.34 | 38.42 | 38.17 | 4.91 | 38.95 | 39.40 | 2.56 |
| BDR-Z | 31.17 | 30.93 | 6.19 | 35.82 | 37.56 | 5.24 | 38.12 | 38.17 | 4.91 | 38.95 | 39.40 | 2.56 |
| SBDR | **28.80** | 29.50 | 5.18 | 33.49 | 34.50 | 6.67 | **34.16** | 34.39 | 5.87 | 36.74 | 36.65 | 2.50 |
| Ours | 29.61 | 29.36 | 4.68 | **32.78** | 33.31 | 4.85 | 35.38 | 36.39 | 4.02 | **35.76** | 35.45 | 2.89 |

The clustering errors by different subspace clustering methods on MNIST database are shown in both Tables VII and VIII. From them, we can see that ABDR largely outperforms the existing subspace clustering methods over almost cases of MNIST dataset. Even if it is relatively unfair to compare ABDR with SBDR, ABDR performs better than SBDR in most cases with less parameter(s) and convex objective. Even if the number of subjects increases gradually, ABDR still outperforms the existing methods including BDR with a large improvement (e.g., 3.19% on ten subjects, 3.04% on nine subjects, and 3.32% on three subjects). Therefore, this experiment on MNIST database demonstrates the effectiveness and superiority BDR on the challenging Handwritten Digit clustering task. Besides, we show the **Z** (**B**) obtained by ABDR and BDR on two datasets with five and ten subjects, respectively, in Figs. 10 and 11. From them, we can observe that ABDR can obtain more exact block diagonal coefficient matrix than BDR even without prefixing the number of subspaces/blocks. Furthermore, we plot the objective function values of (5) with respect to number of iterations on two subsets of five and ten subjects from MNIST dataset in Fig. 12. It can be seen that the objective function monotonously declines and converges after 20 iterations, indicating a fast convergence rate of ABDR.

TABLE IX
CLUSTERING ERROR (%) OF DIFFERENT SETTINGS OF CONVEX FUSED TERMS IN ABDR ON THE EXTENDED YALEB DATABASE

| Algorithm | 2 subjects | | | 3 subjects | | | 5 subjects | | | 8 subjects | | | 10 subjects | | |
|---|---|---|---|---|---|---|---|---|---|---|---|---|---|---|---|
| | mean | median | std | mean | median | std | mean | median | std | mean | median | std | mean | median | std |
| column fusion | 2.11 | 0.78 | 4.67 | 4.04 | 1.56 | 7.46 | 4.33 | 3.75 | 3.33 | 6.22 | 6.25 | 3.75 | 6.39 | 4.06 | 4.70 |
| row fusion | 2.03 | 0.78 | 3.77 | 3.70 | 1.56 | 5.51 | 4.98 | 4.06 | 3.42 | 6.01 | 6.15 | 3.33 | 6.63 | 4.22 | 4.48 |
| ABDR | **0.74** | 0.00 | 1.47 | **2.27** | 0.78 | 5.41 | **3.20** | 2.03 | 2.76 | **2.79** | 2.34 | 2.70 | **4.16** | 2.97 | 3.53 |

TABLE X
CLUSTERING ERROR (%) OF DIFFERENT SETTINGS OF CONVEX FUSED TERMS IN ABDR ON THE MNIST DATABASE

| Algorithm | 6 subjects | | | 7 subjects | | | 8 subjects | | | 9 subjects | | | 10 subjects | | |
|---|---|---|---|---|---|---|---|---|---|---|---|---|---|---|---|
| | mean | median | std | mean | median | std | mean | median | std | mean | median | std | mean | median | std |
| column fusion | 32.64 | 31.58 | 6.74 | 31.85 | 32.07 | 5.49 | 34.64 | 36.38 | 5.72 | 37.22 | 37.39 | 5.31 | 38.41 | 38.95 | 2.83 |
| row fusion | 33.82 | 31.92 | 5.96 | 33.06 | 31.71 | 5.41 | 37.33 | 36.56 | 5.47 | 37.94 | 37.94 | 4.59 | 37.15 | 37.20 | 2.79 |
| ABDR | **32.25** | 32.58 | 5.41 | **29.61** | 29.36 | 4.68 | **32.78** | 33.31 | 4.85 | **35.38** | 36.39 | 4.02 | **35.76** | 35.45 | 2.89 |

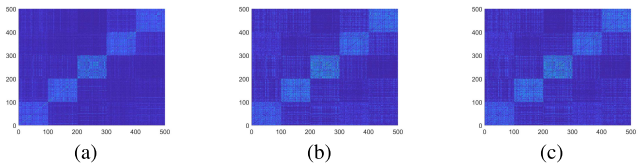

Fig. 10. **Z** (or **B**) obtained by ABDR and BDR on a subset of five subjects from MNIST dataset, respectively. (a) Z obtained by ABDR. (b) B obtained by BDR. (c) Z obtained by BDR.

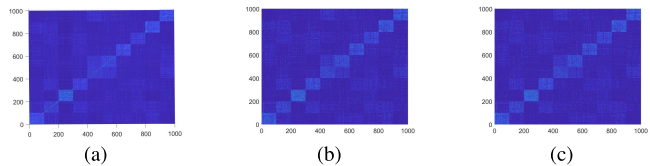

Fig. 11. **Z** (or **B**) obtained by ABDR and BDR on a subset of ten subjects from MNIST dataset, respectively. (a) Z obtained by ABDR. (b) B obtained by BDR. (c) Z obtained by BDR.

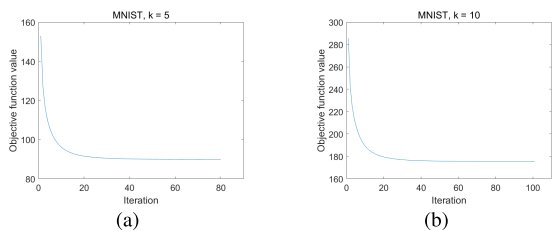

Fig. 12. Objective function value of (5) versus iteration on two subsets from MNIST dataset. (a) Objective function value of (5) versus iteration on a subset of five subjects from MNIST dataset. (b) Objective function value of (5) versus iteration on a subset of ten subjects from MNIST dataset.

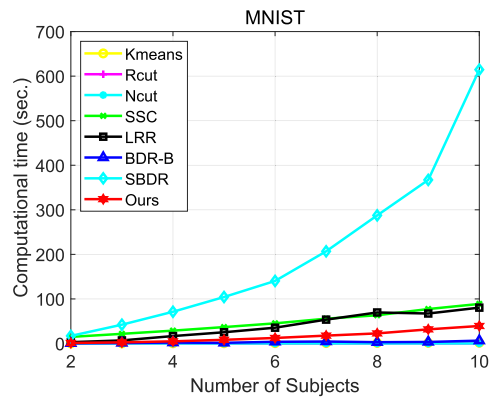

Fig. 13. Average computational time with respect to the number of subjects on MNIST.

Fig. 13 provides the average running time of each method on MNIST with respect to the number of subjects. It can be seen that both ABDR and BDR bring the performance improvement without losing the efficiency. This further validates the low computational complexity of ABDR.

*E. Ablation Study*

To further validate the effectiveness of simultaneous fusion of the columns and the rows in the representation matrix, we delete the column fused term and the row fused term, respectively, and carry out this experiment on Extended YaleB of $k \in \{2, 3, 5, 8, 10\}$ subjects and MNIST of more challenging $k \in \{6, 7, 8, 9, 10\}$ subjects. Tables IX and X show the results. From them, we can find that ABDR with simultaneous fusion of the columns and the rows consistently achieves optimal performance among these three settings, sufficiently verifying the help and effectiveness of simultaneous fusion of the columns and rows in the representation matrix.

*F. Parameter Sensitivity*

In this section, we test the sensitivity of ABDR to hyperparameter settings on real datasets including Extended YaleB with ten subjects, PIE with 68 subjects and MNIST with ten subjects. We show the results in Fig. 14 where the upper row corresponds to $\gamma$ in large range and the lower row corresponds to $\gamma$ lying on the around of the optimal setting. Note that for the convenience of drawing, we set the variable of horizontal axis to be $\log(\gamma)$ in Fig. 14(a)–(c). From Fig. 14(a)–(c), we can see that the clustering performance of ABDR decreases as the hyperparameter $\gamma$ greatly increases. When $\gamma$ is large to a certain extent, ABDR performs steady. This illustrates that when $\gamma$ is larger than the optimal one, the representation coefficients corresponding to the data points from different subspaces begin start to fuse, this actually can help us uncover the relationship between data points from different subspaces. Moreover, comparing Fig. 14(a)–(c) with Fig. 4, we can infer that the low-dimensional subspaces where these real datasets are drawn from intersect and are not strictly independent as the subspaces where the synthetic data points are drawn from. On the other hand, Fig. 14(d)–(f) shows that ABDR performs consistently well when $\gamma$ is relatively small and is not sensitive to the small change of $\gamma$, this makes the selection of optimal hyperparameter $\gamma$ easy.

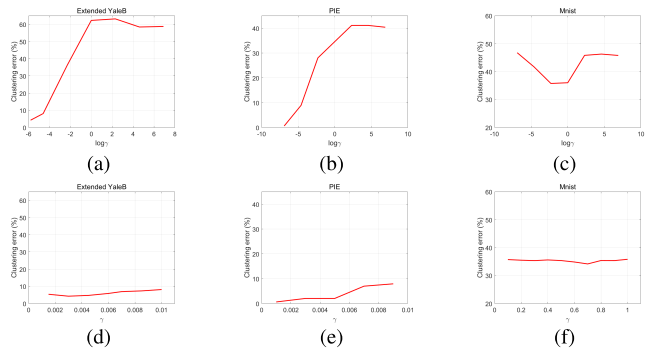

Fig. 14. Clustering error (%) of ABDR as a function of $\gamma$ for ten subjects problems from Extended YaleB, 68 subjects problems from PIE, ten subjects problems from MNIST, respectively. (a)–(c) $\gamma$ in large range. (d)–(f) $\gamma$ lying on the around of the optimal setting.

## V. CONCLUSION

In this article, we focus on an important subclass of the subspace clustering, i.e., spectral-type approaches. Its key first step is to desire learning a representation coefficient matrix with block diagonal structure. So far, two categorical self-expression-based spectral-type subspace clustering methods for achieving such desire, i.e., indirect and direct learning categories, suffer from their respective shortcomings. To compensate their respective shortcomings, inspired by convex biclustering, we follow the direct line and propose a Convex Subspace Clustering method named ABDR. Owing to the convexity of ABDR, the GADMMs can be used to achieve efficient optimization. More importantly, we provide its convergence guarantee. Finally, our experiments on several synthetic data and commonly used benchmarks demonstrate the effectiveness of our proposed ABDR.

There are still some interesting future works, e.g., the theoretical proof of ABDR, extension of the auto-fused terms to tensor subspace clustering and (incomplete) multiview convex subspace clustering [6], [9], deepening ABDR to achieve nonlinear subspace clustering [7], [8], ABDR with discriminative feature learning [60], [61], extension of ABDR with stronger robustness [62], Structured ABDR (SABDR) which incorporates the learning of representation matrix and spectral clustering into a unified framework [53] and so on.

## ACKNOWLEDGMENT

The authors would like to thank all anonymous reviewers for providing valuable suggestions and Chuanxing Geng, Xiang Li for discussion.

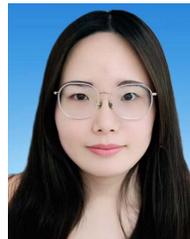

**Yunxia Lin** received the B.S. degree in electronic information engineering from the Qilu University of Technology, Jinan, China, in 2013, and the M.S. degree in communication and information system from Lanzhou University, Lanzhou, China, in 2017. She is currently pursuing the Ph.D. degree with the College of Computer Science and Technology, Nanjing University of Aeronautics and Astronautics, Nanjing, China.

Her research interests include pattern recognition and machine learning.

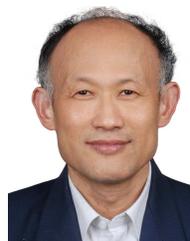

**Songcan Chen** (Member, IEEE) received the B.S. degree in mathematics from Hangzhou University, Hangzhou, China, (now merged into Zhejiang University, Hangzhou), in 1983, the M.S. degree in computer applications from Shanghai Jiaotong University, Shanghai, China, in 1985, and the Ph.D. degree in communication and information systems from the Nanjing University of Aeronautics and Astronautics (NUAA), Nanjing, China, in 1997.

He then worked with NUAA, in January 1986. Since 1998, he has been with the College of Computer Science and Technology, NUAA, as a Full-Time Professor. He has published more than 100 top-tier journals, such as the IEEE TRANSACTIONS ON PATTERN ANALYSIS AND MACHINE INTELLIGENCE, the IEEE TRANSACTIONS ON KNOWLEDGE AND DATA ENGINEERING, the IEEE TRANSACTIONS ON NEURAL NETWORKS AND LEARNING SYSTEMS, the IEEE TRANSACTIONS ON IMAGE PROCESSING, the IEEE TRANSACTIONS ON INFORMATION FORENSICS AND SECURITY, the IEEE TRANSACTIONS ON SYSTEMS, MAN AND CYBERNETICS—PART B, the IEEE TRANSACTIONS ON WIRELESS COMMUNICATIONS and conference papers, such as International Conference on Machine Learning, IEEE Conference on Computer Vision and Pattern Recognition, International Joint Conference on Artificial Intelligence, AAAI Conference on Artificial Intelligence, IEEE International Conference on Data Mining, and so on. His research interests include pattern recognition, machine learning and neural computing.

Dr. Chen is an IAPR Fellow.